\documentclass[sigconf,natbib=false]{acmart}

\usepackage[style=ACM-Reference-Format,backend=bibtex]{biblatex}
\usepackage{booktabs} %

\usepackage{tikz,calc}
\usetikzlibrary{arrows}
\usepackage{csquotes}
\DeclareMathOperator*{\argmax}{argmax}

\usepackage{subcaption}
\usepackage{algorithm}
\usepackage[noend]{algpseudocode}

\setcopyright{rightsretained}

\acmConference[]{}{}{}
\acmYear{2024}
\copyrightyear{2024}

\acmArticle{}
\acmPrice{}

\begin{document}
\title{A Time-Inhomogeneous Markov Model for Resource Availability under Sparse Observations}

\author{Lukas Rottkamp}
\affiliation{%
	\institution{MCML, LMU Munich}
	\streetaddress{Oettingenstr. 67}
	\city{Munich}
	\state{Germany}
	\postcode{D-80538}
}
\email{rottkamp@cip.ifi.lmu.de}

\author{Matthias Schubert}
\affiliation{%
	\institution{MCML, LMU Munich}
	\streetaddress{Oettingenstr. 67}
	\city{Munich}
	\state{Germany}
	\postcode{D-80538}
}
\email{schubert@dbs.ifi.lmu.de}

\begin{abstract}
Accurate spatio-temporal information about the current situation is crucial for smart city applications such as modern routing algorithms. Often, this information describes the state of stationary resources, e.g. the availability of parking bays, charging stations or the amount of people waiting for a vehicle to pick them up near a given location. 
To exploit this kind of information, predicting future states of the monitored resources is often mandatory because a resource might change its state within the time until it is needed. To train an accurate predictive model, it is often not possible to obtain a continuous time series on the state of the resource. For example, the information might be collected from traveling agents visiting the resource with an irregular frequency. Thus, it is necessary to develop methods which work on sparse observations for training and prediction. 
In this paper, we propose time-inhomogeneous discrete Markov models to allow accurate prediction even when the frequency of observation is very rare. Our new model is able to blend recent observations with historic data and also provide useful probabilistic estimates for future states. Since resources availability in a city is typically time-dependent, our Markov model is time-inhomogeneous and cyclic within a predefined time interval. To train our model, we propose a modified Baum-Welch algorithm. Evaluations on real-world datasets of parking bay availability show that our new method indeed yields good results compared to methods being trained on complete data and non-cyclic variants.
\end{abstract}


\maketitle

\section{Introduction}
\label{section:introduction}
Knowledge of a city's resources and the current state of these resources is becoming more and more relevant. For instance, modern routing algorithms integrate this information in order to minimize travel times \cite{ramamohanarao2017traffic, josse2013probabilistic}. This will only get more important when a driver's influence on the route is further diminishing with the advent of autonomous vehicles. For example, the occupation state of parking bays or charging stations, is often required by subsequent algorithms \cite{ayala2011parking}. Other examples for spatial resources are rental cars or returning stations for rental bikes. Though the named examples exhibit varying characteristics, they all have in common that their availability changes over time. 
Apart from routing algorithms, such information can also be used in a augmented reality setting. For example, drivers and pedestrians can be guided to relevant available resources in their vicinity based on visual or auditory clues \cite{losing2014guiding, mata2013augmented}.

But not only the current state of resources is important, it is also often required to estimate the future availability of a resource due to two reasons. First of all, if it is necessary to travel to a resource before claiming it, it is important to have an estimate on the resource availability at arrival time \cite{josse2015probabilistic}. Additionally, having a prediction of the resource availability at a future point in time allows for better travel planning. For example, people might switch to public transportation if the likelihood of claiming a close-by parking bay is low on weekdays. Another application is to help people to estimate the latest time to reserve a resource such as a rental car before the available contingent of resources is claimed by other people.
Having recent information can greatly benefit the prediction, as becomes obvious when looking at extremely short prediction intervals: When the latest available observation of a resource is very recent, the probability of its availability state having changed in the meantime is usually very small. But with the passing of time, this probability increases, until at some point, the latest measurement has lost all its predictive value. For example, the knowledge that a given parking bay was occupied one month ago will have practically no influence on its current state. However, the observed real-time information can be stored to allow the training of general long-term models which describe the average resource usage over time. In this sense, a prediction method for resource availability has to employ a \enquote{hybrid} model, dynamically interpolating between a short-term-model (only considering the latest observation) and a long-term-model (not considering the latest observation at all).

In some cases, real-time information on resources is available, e.g., a city might decide to equip all its parking bays with permanently installed in-ground sensors. For rental vehicles, there often exists a location-based service which allows online tracking of the availability of close-by resources. However, not all types of spatial resources provide gap-less real-time information because maintaining the required sensor network and monitoring system is coupled with considerable expenses. For instance, the majority of on-street parking space around the world is not monitored and many bike sharing services just provide pick-up and drop-off stations but no constant monitoring of available bikes and returning stations. To still provide recent information about resource availability, other approaches have been proposed. 
The \enquote{ParkNet}\cite{mathur2009parknet, mathur2010parknet} project is dedicated to measure parking bay availability based on ultrasonic sensors of passing vehicles. The authors conclude that \enquote{[...] if ParkNet were deployed in city taxicabs, the resulting mobile sensors would provide adequate coverage and be more cost-effective by an estimated factor of roughly 10-15 when compared to a sensor network with a dedicated sensor at every parking space, as is currently being tested in San Francisco.}\cite{mathur2010parknet}. Another method for obtaining the availability state of parking bays is using smartphones: \cite{ma2014updetector} describe multiple indicators for detecting parking and unparking activity. %
To conclude, a prediction method for resource availability should be capable to cope with sparse observations in order to be applicable for a wide range of areas and resources.
A further requirement to model a city's resources is to incorporate time-dependent behavior. For example, a free parking spot in an office district will typically stay vacant much longer during nighttime than a few hours later when employees flock to nearby offices. 
A final aspect which has to be considered is the relation between clustered resources. Since most applications require to claim any available resource but not a particular one, spatially clustered resources should be modeled together. For example, when modeling a station with ten rental bikes, a user is interested if any bike is available the station, but not whether bike-3 is available.

In this paper, we propose a novel prediction method for spatio-temporal resources which fulfills all the requirements named above. Our approach is based on a cyclic time-inhomogeneous discrete Markov model which learns transition possibilities from observed long-term observations but can predict short-term availability of a resource based on the last known observation. Our model is cyclic in time to model the typical change of availability during a certain time period such as a day or a week. Within this period we partition time into a set of discrete steps. For each of these steps, we allow varying transition probabilities which makes our model inhomogeneous in time. To allow the modeling of multiple spatially clustered resources, the states of our Markov model correspond to the number of available resources of a particular location. In order to train our model based on sparse training samples, we allow unknown observations and modify the well-known Baum-Welch expectation-maximization algorithm for hidden Markov models to estimate the parameters of our model. Since the proposed algorithm requires a considerably amount of iterations, we additionally propose a heuristic estimator which requires far less computational effort making the training computable in an embedded system.
To show that our method provides accurate predictions based on the given information, we test our model on the real-world application of modeling parking bays based on two real-world datasets. 

To summarize, this paper's contributions are:
\begin{enumerate}
	\item A discrete, time-inhomogeneous Markov model for predicting the state of a cluster of resources given historic data and (if available) a recent observation of the cluster's state. This model is able to model cyclic behavior, as often occurs in a Smart City setting.
	\item A training algorithm for this model given a observation sequence without missing values.
	\item A second training algorithm based on the Baum-Welch algorithm for hidden Markov models, which shows good performance when only sparse observation sequences are available as training data.
\end{enumerate}

The remainder of the paper is structured as follows: The next section describes existing work regarding the prediction of resource availability, training Markov models on sparse data and other related topics. Then, the problem setting motivated above is formalized in Section~\ref{section:problemsetting}, followed by a definition of  the proposed time-inhomogeneous Markov model. Section~\ref{sec:estimate} details the estimation of model parameters given complete and sparse observation sequences, respectively. This is followed by an evaluation of these algorithms on real-world datasets in Section~\ref{sec:evaluation}. The last section concludes the paper by summing up our achievements and findings, and gives an outlook towards future work.

\section{Related Work}
\label{sec:relatedwork}

Predicting the availability of resources and in particular, parking opportunities in urban environments, is a task with great practical relevance and thus, draws a lot of  attention within the scientific community. \cite{tayade2016advance} surveys a variety of approaches to solve this problem. 

The method being most similar to our approach is \cite{caliskan2007predicting} focusing on the availability of parking bays. The motivation for the work is based on the field of vehicular ad hoc networks (VANETs), i.e. decentralized networks in which agents directly communicate with each other. Since propagating information through such a VANET suffers from significant latency,  information of a parking bay's state might be already outdated  when reaching the agent requiring the information. The authors apply a continuous-time, homogeneous Markov model with constant arrival and parking rates to meet this challenge. In contrast, our approach considers a time-inhomogeneous model which is capable to model the change of availability within the day. Due to our analysis of the used real-world data, this is an important influence because just as traffic density, resource availability might strongly vary over a day.

A further approach modeling resource availability by a continuous-time Markov model is proposed in \cite{josse2015probabilistic}. In this work, the predictions are used to generate a route minimizing the expected time until a free resource is found near the vehicle's destination. The authors also use a time-homogeneous Markov model, i.e. a Markov model whose sojourn times in the two states \enquote{available} and \enquote{consumed} are exponentially distributed with \emph{fixed} parameters defining the means of the respective distribution. Thus, the method also does not consider the time-of-day of the last known observation. As the authors focus on the routing algorithm, the estimation of resources is not the central topic of the paper and therefore, these parameters are not inferred algorithmically from historic data but based on estimates of the authors.

The prediction of resource availability without taking recent knowledge into account is usually understood as the task of solving regression or classification problems with machine learning methods instead of building explicit models. For example, \cite{bock20172} predict the availability of parking bays using Support Vector Regression. The Melbourne dataset also used in our paper's evaluation was used before by \cite{zheng2015parking}, who compare Support Vector Regression, an approach using regression tree and neural networks.%

Above approaches for training Markov models did not focus on finding parameters for training data containing missing values. This was done by \cite{yeh2010estimating} who used an expectation-maximization algorithm and a method using nonlinear equations to find parameters for a discrete homogeneous Markov model. They evaluated their algorithms on a mental health dataset which they reduced to consist of up to $46\%$ of missing data. A different approach was taken by \cite{ma2016bayesian} who train their discrete homogeneous models using a Bayesian approach on data containing up to $25\%$ of unknown data points. The estimation of parameters for homogeneous hidden Markov models trained on data including missing values was examined by \cite{yeh2012intermittent}. They in turn cite \cite{liu1997hidden} as the originator of the idea they used, namely setting the probability of each missing observations to $1$ in the training algorithm.

Note that none of the related work discussed above covers cyclic or otherwise time-dependent Markov models for predicting resource availability. Furthermore, the named methods do not combine short-term and long-term models while at the same time considering time-dependence of the underlying distributions when making predictions.

\section{Problem setting}
\label{section:problemsetting}

We define a \textit{resource cluster} as a group of one or more resources, each of which has a binary state of availability. Each cluster is modeled by one model. Thus, spatially related resources are modeled by the same model. A cluster can be measured at a given point of time $t$, which means that the individual states of its resources become known. The sum of all resources being in the \textit{available} state at this moment is denoted as $O_t$, the observation at time $t$, following the notation of \cite{rabiner1989tutorial}. Without loss of generality, we take the flow of time to be discretized into a sequence of equally-spaced time-steps $t$. Thus, we obtain an observation sequence $O = [O_1,O_2,...,O_m,...]$, in which $O_1$ represents the first and $O_m$ the last observation. In this paper, we examine sparse observations which means that we do not have any observations for most times steps. To represent these missing values, we will denote $O_{t} = -1$ for all time steps $t$ where the resource status is unknown.

Given an observation sequence $O$ of a resource cluster, for example a group of one or more parking bays, our goal is to predict the number of available resources $E({m+d})$ of this cluster at an arbitrary time $d$ after the last observation at time $m$. 

Let us note that in order to predict resource availability for a routing task, we have to train a separate model for each independent cluster of resources. To exploit the spatial relationship between close-by resources, e.g., all bays within the same street, these are modeled by a joint model.

\section{Cyclic Time-Inhomogenous Discrete Markov Model}
\label{section:model}

For modeling the dynamics of resource availability, it is necessary to exhibit time-periodic patterns which suggest using different state transitions probabilities for different points of time. Markov models having this property are called time-inhomogeneous. In order to learn different transition probabilities, we partition time into discrete steps. Only systems having a finite number of discrete states are considered, which must be in exactly one of these states at any given time. Also, as with all discrete Markov models, the state of a model can only change from one step to the next, not during such an interval. In order to describe our new approach, we will start by briefly introducing basic Markov models and then, state our new approach based on cyclic time-inhomogeneous discrete Markov models which is capable to handle sparse observations.

\subsection{Review: Basic Markov Models}
The theory of homogeneous Markov models is well established. \cite{rabiner1989tutorial} provides a good introduction and we will use the notation from this article throughout the paper. The state of a resource cluster at time $t$ is denoted by $q_t$ and assumes one of the $N$ possible states in $S = \{S_1,...,S_N\}$. In our case, $S=\{0,...,M\}$ with each state representing a number of currently available resources in a cluster of size $M$. Typically, a Markov model is specified by a set of states $S$, its state transition probability distribution $A$ and the initial state distribution $\pi$. In discrete models, $A$ is given by a matrix $A = \{a_{ij}\}$ with $a_{ij}$ being the probability of arriving in $S_j$ when taking one time-step forward from $S_i$:
\begin{equation}
\label{eqn:a_homogenous} 
a_{ij} = P[q_{t+1} = S_j | q_t = S_i]
\end{equation}
The initial state distribution is denoted by vector $\pi$ with $\pi_i = P[q_1 = S_i]$, each element indicating the probability of starting a given chain at the respective state.

In the basic model, $q_t$ equals $O_t$, i.e., observations directly reflect the state of the model. The probabilities of being in each of the $N$ states at time $t$ are given by the state probability (row) vector $s_t=(s_t(1),...,s_t(N))$. Let us note that if the state is known, i.e. $q_t = S_i$, the respective entry $s_t(i)$ will be $1$, all others will be $0$.

If we know $s_t$, the most likely state $q_t$ is given by:
\begin{equation}
\label{eqn:q_argmax} 
q_t=\argmax_{1 \le i \le N} s_t(i)
\end{equation}

However, if the set of states is not merely nominal (such as $S=\{\textit{rain}, \textit{snow}, \textit{dry}\}$) but indicates a value on a linear scale, e.g. as in our setting $S=\{0,...,M\}$ with each state representing a number of currently available resources in a resource cluster of size $M$, it is advantageous to not take the state with the highest probability as prediction, but to calculate the expected value over all states:
\begin{equation}
\label{eqn:q_expected} 
E(t)=\sum_{i=1}^{N} S_i s_t(i)
\end{equation}

Our goal is to predict the number of available resources $E({m+d})$ at an arbitrary time $d$ after the last observation $O_m$.  Interpreting the state transition probability distribution $A$ as a matrix, we first calculate the most likely state of the Markov model using Equation~\ref{eqn:s_homogenous} \cite{ross2014introduction} and then use this state to estimate $E({m+d})$.
\begin{equation}
\label{eqn:s_homogenous} 
s_{m+d}=s_{m} \prod_{j=1}^{d} A
\end{equation}
The intuition is that we start at the last known state $s_{m}$ and from that iterate through the chain, at every time-step multiplying the state transition probabilities to obtain the state probability vector $s_{m+d}$ at our target time. Having this, we can use Equation~\ref{eqn:q_expected} to obtain $E({m+d})$.

\subsection{The time-inhomogeneous Cyclic Discrete Markov Model}

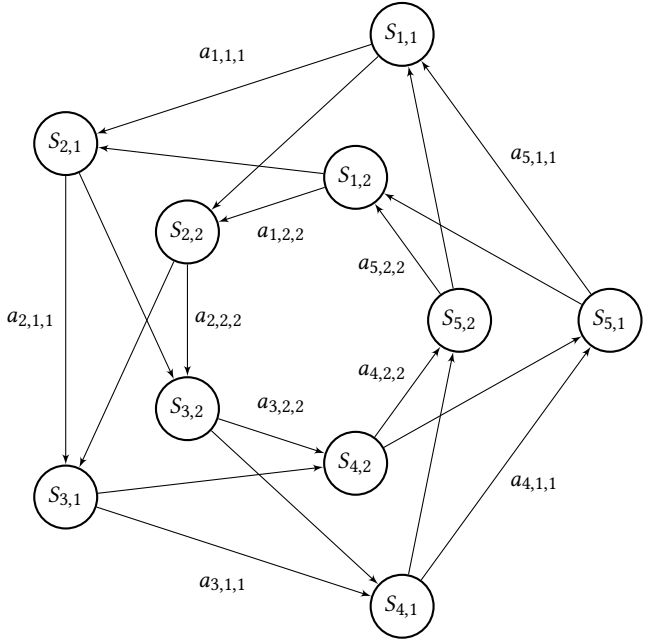
\begin{figure}
	\begin{tikzpicture}
	\tikzset{vertex/.style = {circle, draw, thick, text centered}}
	\tikzset{edge/.style = {->,> = latex'}}
	\foreach \a/\b in {1/1.5,2/2.5,3/3.5,4/4.5,5/5.5}{
		\node[vertex] (A\a) at (\a*360/5: 4cm) {$S_{\a,1}$};
		\node[vertex] (C\a) at (\a*360/5: 2cm) {$S_{\a,2}$};
		\node[] (Z\a) at (\b*360/5: 3.7cm) {$a_{\a,1,1}$};
		\node[] (Z\a) at (\b*360/5: 1.2cm) {$a_{\a,2,2}$};
	}
	\foreach \x/\y in {1/2,2/3,3/4,4/5,5/1} {
		\draw[edge] (A\x) edge node{} (A\y);
		\draw[edge] (C\x) edge node{} (C\y);
		\draw[edge] (A\x) edge node[auto]{} (C\y);
		\draw[edge] (C\x) edge node[auto]{} (A\y);
	}
	\end{tikzpicture}
	\caption{Exemplary visual representation of a cyclic Markov chain with period length $p=5$ and two states. $S_{t,i}$ is used as a shorter notation for $q_t = S_i$. The edges denote the respective transition probabilities. Note that not all edges are labeled as the picture becomes unreadable otherwise. The edges between $S_{i,1}$ and $S_{i+1,2}$ should be labeled with $a_{i,1,2}$, the edges between $S_{i,2}$ and $S_{i+1,1}$ with $a_{i,2,1}$.}
	\label{fig:cyclicMarkovChain}
\end{figure}

For many real-world applications, transition probabilities from one state to another are not constant but time-dependent. Thus, it is necessary to allow varying transition probabilities for different points in time. To be able to learn a proper prediction for a future point in time, we need to make an assumption that the transition probability of the future step can be learned from previous observations. Therefore, we consider the time in our model to follow an infinite cycle of repeating periods. A cycle period might reflect a day or any other reasonable time interval. Furthermore, the last time step of a perdiod is always followed by the first time step of the next period. This cyclic behaviour is observed for many use-cases modelling the ordinary behavior of traffic density, sales numbers or resources as can be seen in our evaluation. Thus, we can learn model parameters for each time step in a period based on the observations for the same times steps in previous cycle periods.

The period length of a cycle is denoted as $p$. As we now have one state transition probability distribution for each possible cycle position $x$, we no longer have a single transition matrix A, but a set of transition matrices $A_x = \{a_{xij}\}$ with $a_{xij}$ with
\begin{equation}
\label{eqn:a_inhomogenous} 
a_{xij} = P_x[q_{t+1} = S_j | q_t = S_i]
\end{equation}
Note that while $t$ may be any positive integer, $x$ runs from $1$ to the period length $p$. The cyclic model's initial state distribution $\pi$ is likewise extended to include the cycle position $x$: $\pi = \{\pi_{xi}\}$ with $\pi_{xi} = P_x[q_1 = S_i]$

Figure~\ref{fig:cyclicMarkovChain} gives an example for such a cyclic Markov chain with $p=5$ and $N=2$. In real-world applications, $p$ will typically be larger: In the evaluations presented below, the length of one cycle was chosen to be one day, with each time-step equaling one minute, resulting in $p=1440$.

\subsubsection{Evaluating the model}

Knowing the state $s_m$ at time $m$, calculating the state probability vector $s_{m+d}$ after $d$ time-steps is straightforward: Equation~\ref{eqn:s_homogenous} has to be modified to use the appropriate state transition matrix $A_x$ when iterating through the chain. Note that $x$ and $t$ are aligned in a way that $q_1$ corresponds the first cycle position ($x=1$). Therefore, the vector $s_{m+d}$  equals:\footnote{Since our indices start with $1$, we have to subtract $1$ before applying the $\bmod$-function, and add $1$ afterwards.}
\begin{equation}
\label{eqn:s_inhomogenous_cyclic} 
s_{m+d}=s_{m} \prod_{j=m}^{m+d-1} A_{((j-1) \bmod{} p)+1}
\end{equation}
As before, we obtain $E({m+d})$ using Equation~\ref{eqn:q_expected}.

It may be useful to look at a concrete example using a simple resource cluster consisting of only one resource: Assume that the current time is 9am sharp. Further assume that the last available observation of the cluster's resource was \textit{available} at 8am (minute-of-day $m=481$)\footnote{Using the equation $\text{minuteOfDay}(t) = 60\text{hour}(t) + \text{minute}(t) + 1$}, i.e. $s_{481}=(0,1)$, and we would like to predict the state at 10am on the same day. Our prediction distance $d$ therefore equals 120 minutes. Now we just multiply forward through the chain using Equation~\ref{eqn:s_inhomogenous_cyclic}, and obtain the state vector indicating probabilities according to our model. Note that it is possible to cycle through the chain multiple times: If one is interested in the probabilities at 10am the next day, the prediction period would be 26h i.e. $d=120+1440$. Further note that the current time is not being used, which makes sense as it gives us no information and therefore does not influence our prediction.

\section{Estimating model parameters for our model}
\label{sec:estimate}

In this paper, model parameters are estimated on historic data from previously observed cycle periods, with the only expert knowledge being the cycle length $p$. This enables an easy transfer between domains: We just have to pick $p$ and supply an appropriate dataset. For estimating parameters, two different scenarios have to be considered. Firstly, if gap-less historic data is available, the Markov model can be parametrized in the established way\cite{ross2014introduction}, which only has to be slightly adapted to reflect the time-inhomogeneity of the state transition probability distribution $A$. This modified algorithm is described in the next subsection. Secondly, if observations are missing, we need to employ another algorithm. In this case, we need to distinguish between observations and states and thus, we employ a modified Baum-Welch algorithm (c.f. Subsection~\ref{subsubsec:estimate_incomplete}). Finally, we will present a very efficient but less accurate algorithm which estimates distribution parameters for sparse data in a single pass.

\subsection{Estimating parameters from complete data}
\label{subsec:estimate_complete}{}

When the observation sequence $O$ and therefore also the state sequence $Q$ of the Markov model is completely known in our observation interval $[t_1,...,t_m]$, i.e. we know $q_t$ at each time $t$ in this interval, we can compute the optimal parameters in a straightforward way\cite{rabiner1989tutorial}: First of all, for each tuple ($t$,$i$,$j$), we calculate the state transition probabilities $P(q_{t+1} = S_j | q_t = S_i, Q)$ according to our data. We obtain these values by counting the following occurrences in our observation sequence. The  probability of being in state $S_i$ at time $t$ and in $S_j$ in the next time-step $P(q_t = S_i, q_{t+1} = S_j | Q)$, divided by the probability of being in state $S_i$ at time $t$ $P(q_t = S_i | Q)$:

\begin{equation}
\label{eqn:train_inhomogenous_complete_probs}
\begin{aligned} 
P(q_{t+1} = S_j | q_t = S_i, Q) &= \frac{P(q_t = S_i, q_{t+1} = S_j | Q)}{P(q_t = S_i | Q)}
\end{aligned}
\end{equation} 

In the homogeneous case, both nominator and denominator would be accumulated over all $t$ for all transitions ($i$,$j$). Thus, the probability distribution for any state is computed by determining the relative portion of expected state transitions for each target state. However, in the cyclic case, we only need to consider $t$ in the set $\theta(x)$ of times belonging to the same cycle position $x$ given by:

\begin{equation}
\label{eqn:train_inhomogenous_theta}
\begin{aligned} 
\theta(x) =\{t \mid 1 \le t < T \wedge ((t-1) \bmod p) + 1 = x \}
\end{aligned}
\end{equation}

The state transition probabilities $a_{xij}$ for the respective cycle positions $x$ and states $i$,$j$, can then be calculated by accumulating and normalizing only the probabilities satisfying this restriction:

\begin{equation}
\begin{aligned}
\label{eqn:train_inhomogenous_complete_as} 
a_{xij} &= P(q_{t+1} = S_j | q_t = S_i, Q,t \in \theta(x))\\
&= \frac{\sum_{t \in \theta(x)} P(q_t = S_i, q_{t+1} = S_j | Q)}{\sum_{t \in \theta(x)} P(q_t = S_i | Q)}
\end{aligned}
\end{equation}

\subsection{Estimating parameters from sparse data}
\label{subsubsec:estimate_incomplete}

If the observation sequence $O$ is not completely known, i.e. at least some observations $O_t$ equal $-1$ , we apply an algorithm based on the Baum-Welch algorithm for hidden Markov models (HMM). This is motivated by the fact that though the states of our model are not hidden in the traditional sense of a HMM (where states are never directly measurable, which they are in our setting), they are hidden in the sense of \enquote{being unknown to us} when not having observations for them. Decoupling actual states from observations will also allow the training on noisy data. For example, if using a camera-based recognition system with accuracy less than 100\%.

To match our model to the HMM notation, we only have to introduce an observation symbol probability distribution $B = {b_j(k)}$ for each $j$ of our $N$ states\cite{rabiner1989tutorial}:
\begin{equation}
\label{eqn:b_inhomogenous} 
b_{j}(k) = P[v_k \textit{ at } t | q_t = S_j]
\end{equation}

Note that in our case, the set of individual symbols $V = \{v_1,... ,v_M\}$ is defined to equal the set of states $S$, so $N=M$. Also note that $B$ is not dependent on the cycle position $x$, as this was not needed for our use-cases. However, this extension can easily be done and may for example be appropriate if a camera-based recognition system principally has a lower accuracy at night. 

To be able to use the Baum-Welch algorithm, we have to adapt it to our cyclic method: The equations for calculating the forward variables  $\alpha$, the backward variables $\beta$, the intermediate variables $\xi$ have to be adjusted from the variants given in \cite{rabiner1989tutorial} to the equations below. The only changes needed are the replacement of $a_{ij}$ with $a_{xij}$, with $x$ being the position in the cycle. As defined above, $x$ is calculated from $t$ using the formula $x=((t-1) \bmod p) +1$, with $p$ being the cycle length. Note that $\alpha$ and $\beta$ are calculated recursively, so for both of them, one equation for initialization and one for subsequent induction steps is given.

\begin{equation}
\label{eqn:alpha_inhomogenous_cyclic} 
\begin{aligned} 
\alpha_{1}(j) &= b_j(O_1) \\ 
\alpha_{t+1}(j) &= b_j(O_{t+1}) \sum_{i=1}^{N} \alpha_t(i)a_{xij}\\
\beta_T(i) &= 1 \\ 
\beta_{t}(i) &= \sum_{j=1}^{N} a_{xij} b_j(O_{t+1}) \beta_{t+1}(j)\\ 
\xi_{t}(i,j) &= \frac{\alpha_t(i) a_{xij} b_j(O_{t+1}) \beta_{t+1}(j)}
{\sum_{i=1}^{N} \sum_{j=1}^{N} \alpha_t(i) a_{xij} b_j(O_{t+1}) \beta_{t+1}(j)}
\end{aligned} 
\end{equation}

The equations for intermediate variables $\gamma$ are not changed, as they do not include state transition probabilities. Finally, the next iteration's state transitions $a_{xij}$ can be determined analogously to Equation~\ref{eqn:train_inhomogenous_complete_as} above:

\begin{equation}
\label{eqn:train_inhomogenous_cyclic_as}
\begin{aligned}
a_{xij} &= P(q_{t+1} = S_j | q_t = S_i, Q,t \in \theta(x),a,b)\\
&= \frac{\sum_{t \in \theta(x)} P(q_t = S_i, q_{t+1} = S_j | O,a,b) }{\sum_{t \in \theta(x)}  P(q_t = S_i | O,a,b) } \\
&= \frac{\sum_{t \in \theta(x)} \xi_t(i,j) }{\sum_{t \in \theta(x)} \gamma_t(i) }
\end{aligned}
\end{equation}

In this paper, the observation symbol probability distribution $b$ is not learned but held constant, nor is it dependent on the time, although these extensions can be made if necessary. We also assume no noise in our measurements, i.e. the observation (if available) always equals the state. But, to cope with missing observations, we apply the \enquote{trick} of setting $b_{j}(-1) = 1$, i.e. assuming that the probability of an unknown observation is always 1\cite{yeh2012intermittent,liu1997hidden}. Looking at the equations above, the intuition behind this is that if we don't know the state at this point in time, all states are possible and therefore also all possible paths containing this state have to be considered when estimating the state transition probability distribution. Thus we arrive at:

\begin{equation}
\label{eqn:b_inhomogenous_fixed} 
b_{j}(k) =
\begin{cases}
1, & \text{if }\ j=k \lor k=-1 \\
0, & \text{otherwise}
\end{cases}
\end{equation}

Given these equations, the estimation of $A_x$ can be done as usual for HMMs: First, each matrix $A_x$ is initialized with suitable values, for example a distribution which assigns the majority of the probability mass to the event of staying in the same state, i.e. the diagonal values are almost one and all other values share the remaining probability mass\footnote{Generally, when implementing these algorithms, it should be made sure that transition probability matrices do not contain zeros, as this may lead to unexpected results such as transition probability distributions not summing to 1 or poor training performance as such transitions then are taken to be \enquote{impossible} and therefore not considered during training.}. Another initialization strategy would be to use the results of another, preferably fast algorithm for estimating $A_x$, such as the algorithm given in Section~\ref{subsec:estimate_heuristic}, which may accelerate the procedure as less iterations are needed until convergence.

Using the initial $A_x$, and the fixed parameters $\pi$ and $b$, values for $\alpha$, $\beta$, $\xi$ and finally $a_{xij}$ are calculated by evaluating the equations given above in this section. These $a_{xij}$ then form the new $A_x$, which is then used for another iteration of this process, and so on, until a convergence criterion is met and the last $A_x$ is returned\cite{rabiner1989tutorial}. We stopped the process when the transition probability distributions changed less than an $\epsilon$ threshold between two subsequent iterations.

Note that due to this iterative nature and the calculation of several intermediate steps needing full passes through the whole observation sequence, the proposed algorithm takes more time than the algorithm presented before in Subsection~\ref{subsec:estimate_complete}.

\subsection{Heuristic Approach for Parameter Estimation on Sparse Data}
\label{subsec:estimate_heuristic}

\begin{figure}
	\begin{tikzpicture}[x=3.7em,y=1.6em]
	\tikzset{vertex/.style = {circle, draw, thick, text centered}}
	\tikzset{edge/.style = {->,> = latex'}}
	\foreach \x/\y in {2/2} {
		\node[vertex] (A1) at (1*\x, 1*\y) {$S_{1,1}$};
		\node[vertex,dashed] (C1) at (1*\x, 0*\y) {$S_{1,2}$};
		\node[vertex] (A2) at (2*2, 1*\y) {$S_{2,1}$};
		\node[vertex] (C2) at (2*\x, 0*\y) {$S_{2,2}$};
		\node[vertex] (A3) at (3*2, 1*\y) {$S_{3,1}$};
		\node[vertex] (C3) at (3*\x, 0*\y) {$S_{3,2}$};
		\node[vertex,dashed] (A4) at (4*2, 1*\y) {$S_{4,1}$};
		\node[vertex] (C4) at (4*\x, 0*\y) {$S_{4,2}$};
	}
	\foreach \a/\b/\x/\y/\z/\u in {1/2/2/0/0/2,2/3/1/1/1/1,3/4/0/2/0/2} {
		\draw[edge] (A\a) edge node[auto]{\x} (A\b);
		\draw[edge] (C\a) edge node[auto]{\y} (C\b);
		\draw[edge] (A\a) edge node[auto]{\z} (C\b);
		\draw[edge] (C\a) edge node[auto]{\u} (A\b);
	}
	\end{tikzpicture}
	\caption{Example of determining edge counts for the training algorithm described in \ref{subsec:estimate_heuristic}. $S_{t,i}$ is used as a shorter notation for $q_t = S_i$. Assume we have observed state $S_1$ at $t=1$ and $S_2$ at $t=4$. There exist four distinct paths from $S_{1,1}$ to $S_{4,2}$. Each edge is annotated with the number of paths visiting it. These counts will be determined for each pair of adjacent observations, combined and normalized to obtain a state transition probability distribution.}
	\label{fig:markovChainAlgorithmIncomplete}
\end{figure}
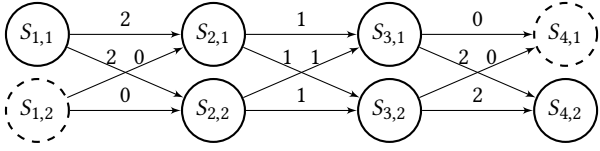

The modified Baum-Welch method presented above has the disadvantage of needing relatively much computing time due to being an iterative method. Therefore, a faster alternative has been devised, which only needs one pass through the observation sequence.

Our goal is to make reasonable guesses about what happened during each gap in the observation sequence, i.e. each streak of time-steps for which no information is available. As an example, consider that we observed $q_1 = S_1$ and $q_4 = S_2$, but not the states in between, as visualized in Figure~ \ref{fig:markovChainAlgorithmIncomplete}. For each such gap, we assume that all paths satisfying a reasonable set of conditions are assigned the same probability of being the path which actually happened. This reflects the fact that without previous knowledge, each of these paths could well be the unobserved path. The remaining paths are assigned a probability of zero. The first, obvious condition is that the path must start and end at the states that have been observed. The second condition is that paths must only include the two measured states and those in-between\footnote{Remember that our states represent the number of available resources in each resource cluster, e.g. parking bays in a larger parking area, an as such are inherently ordered.}. While this second condition may be violated by the true path, it allows for pruning the majority of paths with low likelihood. We choose this condition in favor of other heuristics as it showed good results in preliminary experiments.
Now, for each transition between two states, i.e. each edge in the graph, the number of paths visiting this edge is calculated. These counts are then normalized to form a transition probability distribution for the respective pair as before in Equation~\ref{eqn:train_inhomogenous_complete_as}. This is repeated for each adjacent pair of observations in the observation sequence. The resulting distributions are multiplied, to obtain the final transition probability distribution. The whole algorithm is shown in Algorithm \ref{algo:heuristic}. Note that the resulting state transition probability distribution can be used in exactly the same way as the distributions estimated by the methods described above.

\begin{algorithm}
	\caption{Heuristic for estimating transition probability distributions when having sparse observations}\label{euclid}
	\label{algo:heuristic}
	\begin{algorithmic}[1]
		\Procedure{EstimateStateTransitionProbabilities}{}
		\For{\text{each pair of two adjacent, known observations}}
		\State $\textit{paths} \gets $\parbox[t]{.2\linewidth}{%
			$\text{all paths between those two observations}$ \\
			$\text{which satisfy the conditions given in Sec.~\ref{subsec:estimate_heuristic}}$\;}
		\For{\text{each \textit{edge} visited by at least one path in \textit{paths}}}
		\State $\textit{edgeCt[edge]} \gets \text{number of \textit{paths} visiting this \textit{edge}}$
		\EndFor
		\State $\textit{gapDists[gap]} \gets $\parbox[t]{.2\linewidth}{%
			$\text{calculate state transition probability} $ \\
			$\text{distribution given \textit{edgeCt}}$\;}
		\EndFor
		\State $\textit{combinedDist} \gets $\parbox[t]{.2\linewidth}{%
			$\text{combine distributions in \textit{gapDists} by}$ \\
			$\text{summing their individual probabilities}$ \\
			$\text{and renormalizing the result}$\;}
		\State \Return $\textit{combinedDist}$
		\EndProcedure
	\end{algorithmic}
\end{algorithm}

\section{Evaluation}
\label{sec:evaluation}

\subsection{Implementation}
The algorithms described above were implemented using Java 1.8. As calculated probabilities occasionally become too small for standard data-types as \textit{Double}, calculations are conducted in log-space to achieve numerical stability\cite{mann2006numerically}. Our algorithms were also compared with an SVM for regression included in the publicly available LibSVM package for Java. LibSVM is a widely used standard implementation of various SVM procedures \cite{CC01a}. It is released under the modified BSD license.

\subsection{Data used for evaluation}
\label{subsection:evaluation_data}

For evaluating our algorithms, real-world datasets are used. Both the Australian Capital Territory (ACT) Government\footnote{Australian Capital Territory (ACT) open data platform: https://www.data.act.gov.au; Data licensed under \enquote{Creative Commons Attribution 4.0 International}: https://creativecommons.org/licenses/by/4.0} and the City of Melbourne\footnote{City of Melbourne open data platform: https://data.melbourne.vic.gov.au; Data licensed under \enquote{Creative Commons Attribution 3.0 Australia}: https://creativecommons.org/licenses/by/3.0/au/}, Australia, host open data platforms on which historic data of parking bay occupations is provided under creative commons license. These data were monitored through permanently installed in-ground sensors for every second, and are therefore deemed to be very accurate.

The datasets contain \enquote{stays}, each representing a parking event consisting of an arrival and departure time. \enquote{Stays} with a duration larger than 24h day were excluded, as they typically indicated a corrupt measurement.

\subsubsection{Dataset Canberra}
This dataset was constructed from the dataset \enquote{Smart Parking Stays}\footnote{https://www.data.act.gov.au/Transport/Smart-Parking-Stays/3vsj-zpk7}, recorded by the ACT government during their \enquote{SmartParking} trial and covers the time from January 2016 till end of July 2017. Over this time, parking events for 454 parking bays in the Manuka district of Canberra were observed, resulting in 3.72 million records, each containing a reference to the recorded bay, the arrival and departure time for this stay, and meta data such as allowed parking duration.

Of the original 454 parking bays, only \enquote{on-street} bays were considered, i.e. bays which do not belong to separated parking lots as these do not directly apply to our eventual use-case of recording the occupation state of bays using agents participating in normal traffic. Further, bays with exceptional permissions were excluded, such as short-term (less then 15 minutes) parking opportunities or bays for the disabled. This left 160 bays, which were grouped into 21 resource clusters by their respective \enquote{LotCode} given in the original dataset. Thus, the individual resources are grouped by their spatial closeness: Bays in the same resource cluster are directly adjacent to another and therefore are expected to have similar availability distributions.

\subsubsection{Dataset Melbourne}
The \textit{Melbourne} dataset was built using the dataset \enquote{Parking bay arrivals and departures 2014}\footnote{https://data.melbourne.vic.gov.au/Transport-Movement/Parking-bay-arrivals-and-departures-2014/mq3i-cbxd}, provided by the city of Melbourne. It contains data from the beginning of January 2014 to the end of December 2014. In total, stays for 5228 distinct bays are recorded, totaling in 13.5 million records. Each record again contains a reference to the bay in question, arrival and departure time, and various other information.

Of the 5228 bays, six contained too little data, and therefore were not considered, leaving 5222 bays in the dataset.

The dataset includes information about the bays' location including their street and neighboring intersections. From this, street segments were inferred, each street segment being a streak between two intersections. Our application considers the recording of availability information by visiting agents, which may not be able to record parking bays' state across the street due to obstructions. Thus, the side of the road was also considered when building street segments, i.e. a paved area between two intersections consists of two street segments. Bays were grouped into resource clusters by the street segment they are located in, resulting in 291 clusters.

\subsection{Experiments}
\label{subsec:evaluation_experiments}

For evaluation, distinct training and testing sets were built. It was determined to take two weeks of data for training, to give the algorithms sufficient data to work with while also allowing use-cases where not much data is available. The testing data spans six weeks, so that eventual outliers in this set are leveled out. To minimize possible concept drift, all weeks were adjacent and the training weeks were surrounded by testing data: Weeks three and six of a total of eight weeks were taken as the training set $Tr_{0}$, while the remainder forms the testing set $Te$.

Only weekdays were considered for both training and testing, to simplify evaluation: Because parking distributions of weekdays and weekends greatly differ, building joint models was not an option. It was determined by the authors that the constriction to weekdays is suitable for the evaluation of the presented models and their training methods. As cycle period was chosen to be one day, i.e. each weekday was assumed to show roughly the same behavior as is supported by the data.

To evaluate the algorithms under sparse data, additional training sets were constructed from the gap-less training set $Tr_{0}$. As gaps appear when no agent with appropriate measuring equipment visits the street segment in question at that time, this mechanism was approximated to generate reduced datasets. For simplicity, it was assumed that the arrivals of measuring agents are independent and occur with a constant rate over the day. Thus, the time between two subsequent arrivals is exponentially distributed with arrival rate $\lambda$. The mean of these times is given by $\beta = \lambda^{-1}$ and will be used as our parameter for sampling \cite{ross2014introduction}: The sizes of individual gaps in the set $Tr_{\beta}$ are randomly drawn according to the exponential distribution with mean $\beta$.

We compare the following prediction methods: Our modified Baum-Welch algorithm (\textbf{BW}) and the heuristic (\textbf{HEUR}) described in Section~\ref{sec:estimate} are compared to the standard algorithm for gap-free data (\textbf{STD}), to be able to see if our methods are justified at all. For further comparison, the trivial algorithm of always \enquote{predicting} the last measured value (\textbf{LAST}) was included. This algorithm is expected to perform very good for short-term predictions as changes of resource availability happen over time. The fifth model always returns the historic average (\textbf{AVG}) for the minute-of-day in question, not taking the last measurement into account at all. In this way, \textbf{AVG} represents the complement to \textbf{LAST} because it only depends on historic data but does not consider the last known state. Last but not least, we compare our method to SVM regression  (\textbf{SVM}) \cite{bock20172} which represents a state-of-the-art approach for long-term predictions. For each resource cluster, one model was trained for each training set. Note that all models for each resource cluster were trained and tested independently from the other clusters.

All results were averaged over four repetitions, to reduce the variation introduced by the random sampling process, although it should be noted that the sampled training sets of a given $\beta$ were the same for all models during each repetition, to obtain fair scores in this regard. 

\subsubsection{Qualitative comparison of models}

\begin{figure*}
	\centering
	\begin{subfigure}[b]{0.5\textwidth}
		\includegraphics[width=\textwidth]{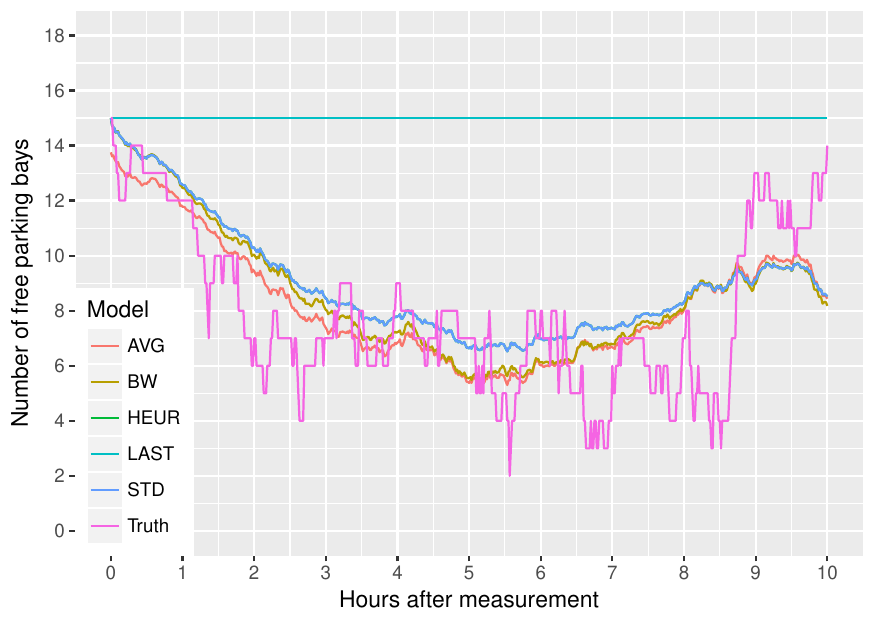}
		\caption{Predictions when trained on complete data.}
		\label{fig:prediction_resource_0}
	\end{subfigure}
	~ %
	\begin{subfigure}[b]{0.5\textwidth}
		\includegraphics[width=\textwidth]{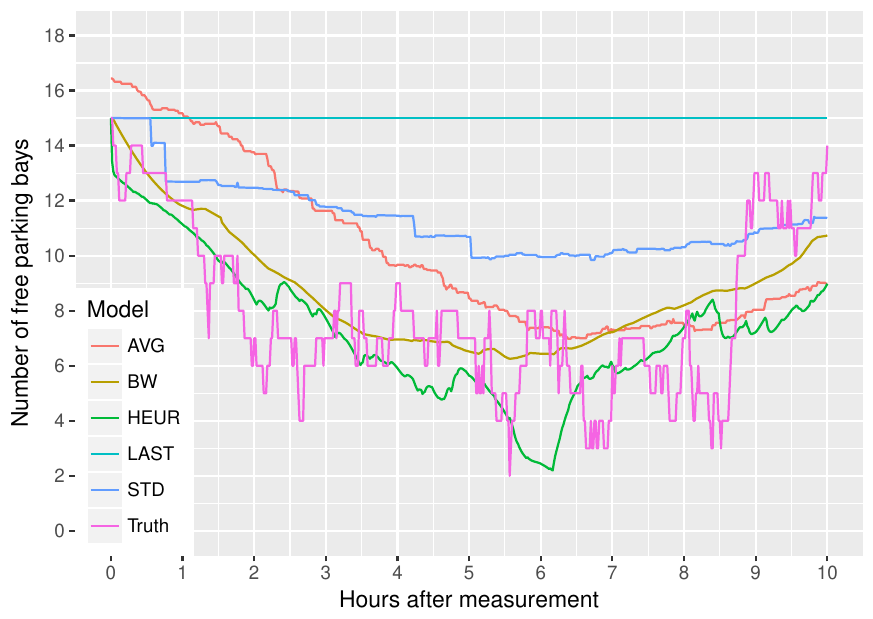}
		\caption{Predictions when trained on sparse data ($\beta=120$).}
		\label{fig:prediction_resource_120}
	\end{subfigure}
	\caption{Predictions and true parking availability for the next 10 hours after a measurement in the morning, for one resource in the Melbourne testing set. Colors of models are the same in all graphs. The algorithms \textbf{HEUR} and \textbf{STD} yield identical models when trained on complete data, thus the graph of \textbf{HEUR} is completely covered by the graph of \textbf{STD}.}
	\label{fig:prediction_resources}
\end{figure*}

Before looking at summarized results, it is insightful to take a look at individual predictions. Figure \ref{fig:prediction_resources} shows predictions for a single day in the test set, when the last measurement of the resource cluster in question was recorded at 8am. Figure \ref{fig:prediction_resource_0} shows results for models trained on complete data. Note that the average \textbf{AVG} starts a little below the actual measurement of fifteen free bays which shows that the cluster was less occupied at this day and time than it was expected given the training data, although in this case the deviation is very small. Still it can be seen how the models exploit this by predicting slightly higher values in the beginning than would be expected if this short-term knowledge would not have been available. It can also be seen that the methods converge to the average as time passes, which can be interpreted as the decreasing value of this short-term knowledge due to noise introduced by the progress of time. In all, the graphs stay fairly similar and no large differences in predicting performance are to be expected. Figure~\ref{fig:prediction_resource_120} presents a completely different picture: Here, models were trained on data with a $\beta$ of 120, i.e. less than $99\%$ of all data points were taken for training. It appears that the average is significantly distorted and also \textbf{STD} does not grasp the trend we saw when trained on complete data. The algorithms \textbf{HEUR} and \textbf{BW} seem to have used the sparse data to much better success, although they look coarser than before. Judging by these graphs, the performance of those algorithms appears to be superior to \textbf{STD}, in this single example.

\subsubsection{Quantitative comparison of models}

Of course the qualitative analysis above is not suited to make general statements about performance, so an extensive quantitative analysis was conducted: Models were trained on datasets of different levels of sparsity ($\beta \in \{30,60,120\}$). Then, each minute in the testing data between 7am and 11pm\footnote{At night, almost all bays are free, making these intervals quite uninteresting.} was taken as a target and predicted by each model given a measurement $d \in \{15,30,60,120,240\}$ minutes before. The individual errors were accumulated per model to obtain mean absolute error (MAE) values. The root mean square errors (RMSE) were also calculated but are not shown here as they provide no further insights and the MAE is more easily interpreted. Because the number of resources differs between resource clusters, these error values are then normalized by the total number of resources of the respective cluster to allow the comparison of clusters of different size: For example, a model evaluating to a normalized MAE of $0.1$ has an absolute MAE of $2.0$ if it represents a cluster of size twenty, i.e. on average the prediction is $2.0$ bays off the real value.

\begin{figure*}
	\centering
	\begin{subfigure}[b]{0.33\textwidth}
		\includegraphics[width=\linewidth]{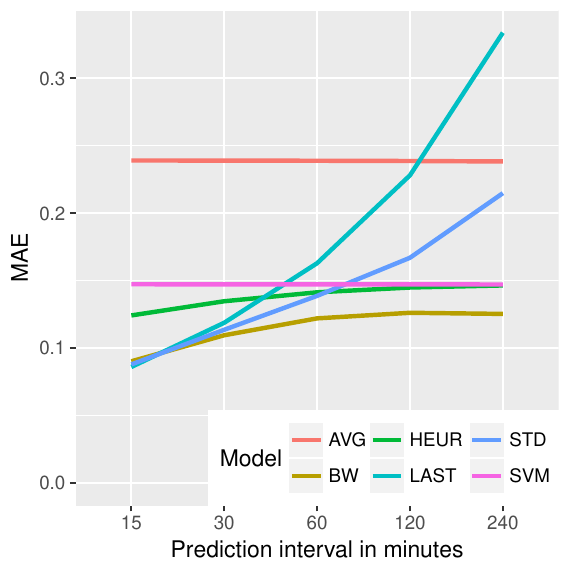}
		\caption{Trained on data sampled with $\beta$ of 30.}
		\label{fig:results1_melbourne}
	\end{subfigure}
	~ %
	\begin{subfigure}[b]{0.33\textwidth}
		\includegraphics[width=\linewidth]{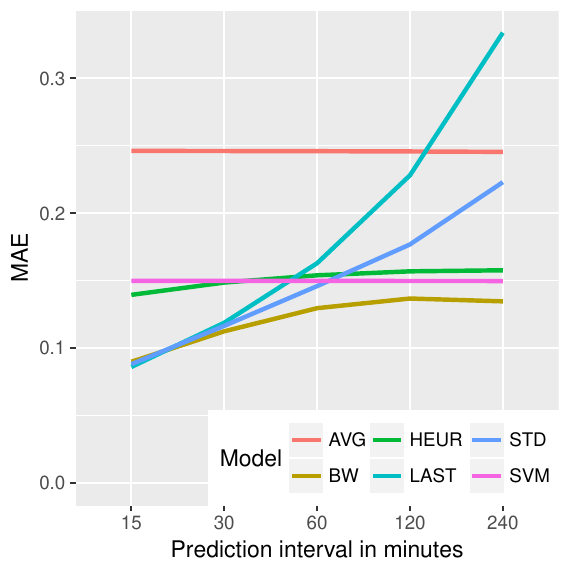}
		\caption{Trained on data sampled with $\beta$ of 60.}
		\label{fig:results1_melbourne_60}
	\end{subfigure}
	~ %
	\begin{subfigure}[b]{0.33\textwidth}
	\includegraphics[width=\linewidth]{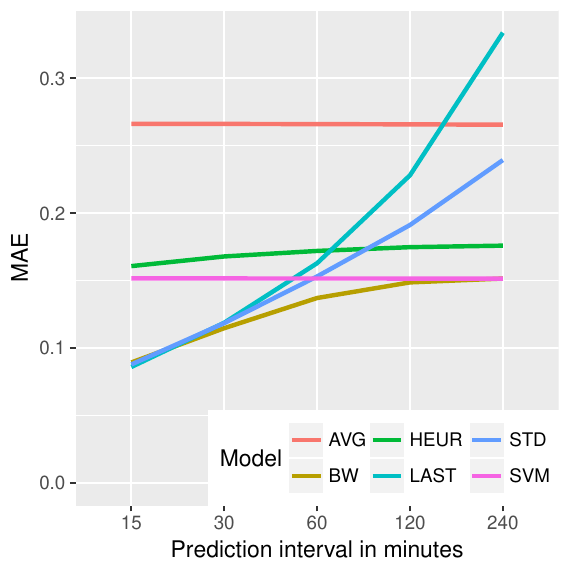}
	\caption{Trained on data sampled with $\beta$ of 120.}
	\label{fig:results1_melbourne_120}
	\end{subfigure}
	\caption{Mean absolute errors (MAE) of the different models for resource clusters containing 5 to 20 parking bays (Melbourne dataset). The MAE is normalized by the number of bays, so a value of 0.1 corresponds to an absolute mean error of 2.0 for a cluster containing 20 bays. These graphs show that the algorithm \textbf{BW} is able to take advantage of knowing the recent measurement at short prediction intervals, but also is able to predict longer intervals, especially if data is not too sparse.}
	\label{fig:results1}
\end{figure*}

\begin{table}
	\caption{Normalized MAE for different models. Models were trained for clusters containing 5 to 20 bays.}
	\begin{tabular}{l|cc|cc}
		\toprule
		& \multicolumn{2}{c}{Canberra} & \multicolumn{2}{c}{Melbourne}\\
		Model & hom. & inhom. & hom. & inhom.\\
		\midrule
			BW & \textbf{0.197} & \textbf{0.142} & \textbf{0.174} & \textbf{0.121} \\
			SVM & -- & 0.172 & -- & 0.150 \\
			HEUR & 0.407 & 0.196 & 0.279 & 0.153 \\
			STD & 0.202 & 0.167 & 0.175 & 0.151 \\
			AVG & 0.423 & 0.348 & 0.315 & 0.250 \\
			LAST & 0.206 & 0.206 & 0.186 & 0.186 \\
		\bottomrule
	\end{tabular}
	\label{tab:results_groups}
\end{table}

Results for the time-inhomogeneous models trained on the Melbourne dataset are displayed in Figure~\ref{fig:results1}: For short prediction intervals as 15 and 30 minutes, the \textbf{BW} model is able to exploit the short-term information given by the last measurement. At 60 minutes, enough time has passed to introduce enough noise from parking and unparking vehicles so that this last measurement is not longer sufficient: The error of \textbf{LAST} increases. Now the long-term model gets more important, as can be seen by \textbf{SVM} becoming competitive. Note that the error of this model is constant as no short-term information are given to this model. When comparing the three figures, it can also be seen that its error is only slowly increasing when data becomes even sparser, i.e. $\beta$ grows. Model \textbf{BW} appears to be more sensitive to this as its advantage shrinks with growing $\beta$.

Accumulated results are shown in Table~\ref{tab:results_groups}, which also includes results for the time-homogeneous counterparts of the models, i.e. models using the same algorithm but not taking the time-of-day into account. Note that the model \textbf{LAST} always ignores the time-of-day by definition, so the results of the time-homogeneous and time-inhomogeneous variants are the same, as would be expected. No time-homogeneous version of SVM was evaluated as it would just output a constant prediction as the latest observation was not included in its training. When comparing the respective values in the table, it becomes obvious that the time-inhomogeneous models show much better performance than their counterparts. This supports our choice to introduce time-inhomogeneous variants. Furthermore, predictions given by the model \textbf{BW} consistently displayed the lowest error values. Our \textbf{HEUR} shows slightly worse performance than \textbf{SVM} at the benefit of being faster, which can be seen later in this section.

\begin{table}
	\caption{Normalized MAE for different models. Models were trained for clusters containing only one parking bay.}
	\begin{tabular}{l|cc|cc}
		\toprule
		& \multicolumn{2}{c}{Canberra} & \multicolumn{2}{c}{Melbourne}\\
		Model & hom. & inhom. & hom. & inhom.\\
		\midrule
		BW & 0.341 & \textbf{0.258} & 0.291 & \textbf{0.230} \\
		SVM & -- & 0.277 & -- & 0.278 \\
		HEUR & 0.568 & 0.462 & 0.391 & 0.334 \\
		STD & 0.331 & 0.293 & 0.271 & 0.241 \\
		AVG & 0.544 & 0.475 & 0.385 & 0.350 \\
		LAST & \textbf{0.273} & 0.273 & \textbf{0.255} & 0.255 \\
		\bottomrule
	\end{tabular}
	\label{tab:results_resourceswithonlyonebay}
\end{table}

To evaluate whether our grouping of spatially similar parking bays into resource clusters and jointly modeling them by one model for each of those clusters is warranted, another experiment was conducted: In this experiment, each resource cluster only contains a single parking bay. Then a single model is trained on each of these resources and evaluated in the same way as before. The resulting MAE values are shown in Table~\ref{tab:results_resourceswithonlyonebay}. Comparing these values to the results for resource clusters spanning spatially related resources (Table~\ref{tab:results_groups}), makes it obvious that combining similar bays improves predictive performance.

\begin{figure}
	\includegraphics[width=\linewidth]{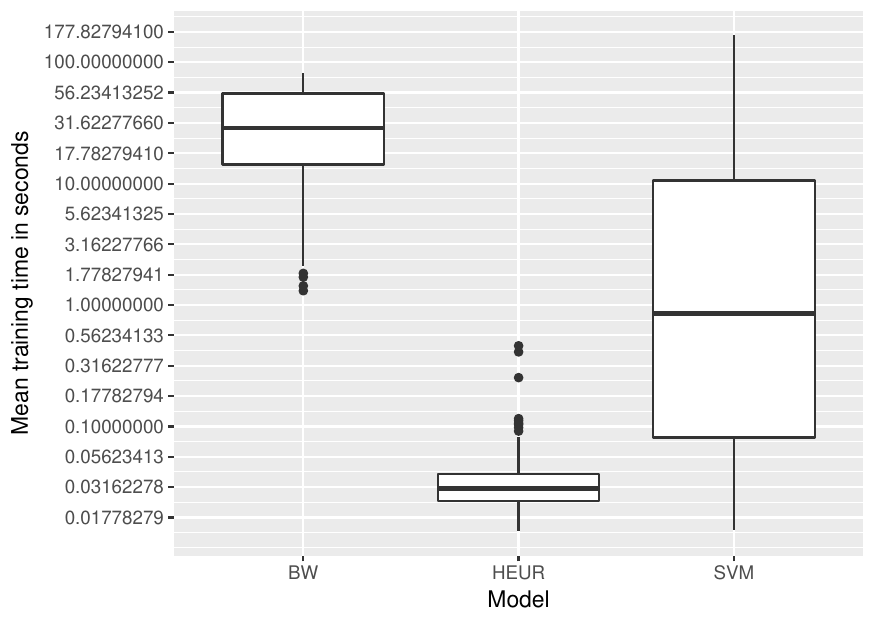}
	\caption{Average runtimes of algorithms for the evaluations shown in Figure~\ref{fig:results1}. Note that the vertical axis is logarithmic.}
	\label{fig:performance_comparison}
\end{figure}

In another experiment, it has been evaluated how much faster our heuristic is. The benchmarks were run on a Intel Core~i7 7700HQ CPU. Note that the training process was parallelized using all logical processors. Each training of a model used a single thread. As training took several hours and the different types of models were interleaved, it can be assumed that potential side-effects e.g. due to the scheduler or operating system did not significantly influence the results. We measured the runtime for each of the 4932 training runs needed for the results shown in Figure~\ref{fig:results1}. Averaged runtimes per training are visualized in Figure~\ref{fig:performance_comparison}. The exact averages are 25.4s for \textbf{BW}, 12.1s for \textbf{SVM} and 0.0388s for \textbf{HEUR}. Thus, \textbf{HEUR} was faster than \textbf{BW} by a factor of 654, and faster than \textbf{SVM} by a factor of 313. The large variance of results for \textbf{SVM} is due to the learning algorithm generally taking longer the more training examples are supplied. This behavior is not shared by the \textbf{BW} algorithm, as it always uses an observation sequence of the same length, in which only the number of \textit{undefined} values varies. On the other hand, runtime of \textbf{BW} depends on the number of resources summarized in one model, as the matrix operations then need to operate on larger matrices. This is not an issue for \textbf{SVM}, making this algorithm a viable alternative if models need to incorporate a large number of resources.

\section{Summary and Outlook}

In this paper, we investigate the problem of predicting the availability state of resources given historic data and a recent measurement. Resources in Smart City settings are usually depending on the time of day. To address this, we propose a time-inhomogeneous Markov model able to model such cyclic behavior. As the historic data typically available in our setting is sparse, we present algorithms designed to cope with sparse data. Because we train our proposed model using a modified Baum-Welch algorithm which needs a number of passes through the observation sequence, we also propose a faster alternative which can be used to shorten the training time while sacrificing prediction performance.

Evaluations on real-world datasets show that the proposed time-inhomogeneous Markov model combined with a Baum-Welch training algorithm yields better predictions than other approaches, as the standard Markov model or an SVM. Our experiments also show that our time-inhomogeneous apporach is warranted, as results are consistently better that time-homogeneous variants of our models and training algorithms, such as the standard Markov model. Further, experiments confirm that time-inhomogeneous models have a better performance than their time-homogeneous variants. The heuristic alternative is proven to need significantly less runtime, making it indeed a useful alternative to our Baum-Welch variant. Evaluations also show that it is favorable to predict spatially related resources using a joint model, instead of building isolated models for each individual resource.

In future work, we want to consider more complex dependencies between nearby or otherwise similar resources in order to challenge the problem of data sparsity even more. This could be achieved through a hierarchy of models which represents the spatial structure of a city. Evaluation methods would then not only blend short-term and long-term observations as in this paper, but also include information of different hierarchical layers.

\begin{acks}
We thank the Australian Capital Territory and the City of Melbourne, Australia, for providing the parking datasets used in this paper under Creative Commons license, as this enabled us (among others) to evaluate our methods on real-world data.
\end{acks}

\setlength{\emergencystretch}{1em} %
\printbibliography

\end{document}